\documentclass[11pt]{article}

\usepackage[preprint]{acl}
\usepackage{amssymb}
\usepackage{amsmath}
\usepackage{booktabs}
\usepackage[most]{tcolorbox}
\usepackage{times}
\usepackage{latexsym}
\usepackage[T1]{fontenc}

\usepackage{threeparttable}

\usepackage{microtype}

\usepackage{inconsolata}
\usepackage{stfloats}

\usepackage{graphicx}
\usepackage{subcaption}
\title{Beyond Overlap Metrics: Rewarding Reasoning and Preferences for Faithful Multi-Role Dialogue Summarization}

\author{
\bfseries Xiaoyong Mei\textsuperscript{1}\thanks{Equal contribution.},
Tingting Zuo\textsuperscript{1}\footnotemark[1],
Da Chen\textsuperscript{2}\footnotemark[1],
Guangyu Hu\textsuperscript{3},
Xiangyu Wen\textsuperscript{4} \\
\bfseries Chao Duan\textsuperscript{1}\thanks{Corresponding author.},
Mingyan Zhang\textsuperscript{1},
Fudan Zheng\textsuperscript{5} \\
\textsuperscript{1}Zhejiang Normal University \quad
\textsuperscript{2}Huawei Technologies \quad
\textsuperscript{3}HKUST, Hong Kong \\
\textsuperscript{4}CUHK, Hong Kong \quad
\textsuperscript{5}Sun Yat-sen University \\
\texttt{\{cdmxy, mingyanzhang, duanchao, zuotingting\}@zjnu.edu.cn} \\
\texttt{915102610106@njust.edu.cn \quad ghuae@connect.ust.hk} \\
\texttt{1155186676@link.cuhk.edu.hk \quad zhengfd5@mail.sysu.edu.cn}
}
  
\begin{document}
\begin{sloppypar}

\maketitle

\begin{abstract}
Multi-role dialogue summarization requires modeling complex interactions among multiple speakers while preserving role-specific information and factual consistency. However, most existing methods optimize for automatic metrics such as ROUGE and BERTScore, which favor surface-level imitation of references rather than genuine gains in faithfulness or alignment with human preferences. We propose a novel framework that couples explicit cognitive-style reasoning with reward-based optimization for multi-role dialogue summarization. Our method first distills structured reasoning traces (e.g., step-by-step inferences and intermediate reflections) from a large teacher model and uses them as auxiliary supervision to initialize a reasoning-aware summarizer via staged supervised fine-tuning. It then applies GRPO with a dual-principle reward that blends metric-based signals with human-aligned criteria targeting key information coverage, implicit inference, factual faithfulness, and conciseness. Experiments on multilingual multi-role dialogue benchmarks show that our method matches strong baselines on ROUGE and BERTScore. Specifically, results on CSDS confirm the framework's stability in semantic consistency, while in-depth analysis on SAMSum demonstrates clear gains in factual faithfulness and model-based preference alignment. These findings underscore the value of reasoning-aware and preference-aware training for reliable dialogue summarization\footnote{Checkpoints and datasets are available at \url{https://huggingface.co/collections/NebulaPixel/summorchestra-multirole-summary}.}.

\end{abstract}

\section{Introduction}

In this paper, we focus on the multi-role dialogue summarization tasks, where the system distills key information from multi-turn conversations into concise, coherent summaries~\citep{Feng2022}. Such summaries enable users to quickly grasp the main points without having to navigate complex conversational context~\citep{chen-yang-2020-multi}. This capability is crucial for a range of real-world applications, including customer service interaction analysis and automatic generation of meeting minutes~\citep{Feigenblat2021,Zhao2021}.

A central challenge in dialogue summarization is accurately modeling complex interaction dynamics to produce faithful summaries. Early work established baselines by fine-tuning pre-trained sequence-to-sequence models~\citep{Lewis2020,Zhang2020,Zhong2022}. However, these methods largely operate as surface-level text mappers and lack deep semantic understanding. More recent research has turned to Large Language Models in an effort to overcome these limitations, considering their impressive performance across a wide range of tasks and application scenarios~\citep{Wang2023,Tian2024,Zhang2025,Wen2025GuidelineCI}. Existing LLM-based approaches primarily seek to elicit reasoning capabilities through Chain-of-Thought prompting~\citep{Wang2023,Jin2025}, or to improve alignment via instruction tuning and Reinforcement Learning from Human Feedback (RLHF)~\citep{Rafailov2023DirectPO,Li2025,Lu2025,Ye2025a}.

Despite these advances, maintaining faithfulness remains a major bottleneck. In particular, we observe that models are prone to hallucinations—such as misattributing responsibility to a speaker or fabricating unsupported details—which severely undermines factual accuracy and alignment with human preferences. A key reason is that real-world conversations are characterized by multiple participants, informal and noisy language, and intricate interaction patterns~\citep{Ramprasad2024,Wen2025GuidelineCI}, but standard instruction tuning does not enforce strict logical consistency, all of which can cause models to miss critical information or misinterpret speaker intent, ultimately leading to factual inconsistencies~\citep{Ramprasad2024}.

To address these challenges, we introduce a novel framework for multi-role dialogue summarization that explicitly embeds cognitive-style reasoning into the reinforcement learning process. Our method is designed around the hypothesis that, before producing a summary, the model should engage in structured reasoning procedures that mirror human cognitive behaviors—such as inferring speaker intent, tracking responsibilities, and verifying factual consistency. By orchestrating this reasoning phase prior to generation, our framework aims to substantially improve both factual faithfulness and alignment with human preferences in complex, real-world conversational settings. 

Specifically, we adopt a teacher–student paradigm~\citep{Hinton2015DistillingTK,Wen2025ReasoningSD,Fang2025KnowledgeDA} to distill structured reasoning capabilities from a larger model into a base model, initializing a reasoning-aware summarizer via supervised fine-tuning. Building on this, we introduce a dual-principle reward mechanism within a group relative policy optimization (GRPO) training framework~\citep{Shao2024}. A \textbf{reasoning principle} evaluates intermediate reasoning traces and summaries in terms of key information coverage, implicit understanding, and factual fidelity, while a \textbf{summary principle} assesses final outputs using lexical and semantic metrics (e.g., ROUGE, BERTScore) and length control. Together, these rewards provide fine-grained process supervision, enabling the policy model to internalize complete reasoning paths and generate summaries that are both faithful and well aligned with human preferences.

Our experiments demonstrate that our method attains performance on par with strong baselines in standard automatic metrics such as ROUGE and BERTScore, while delivering substantial gains in factual faithfulness and preference alignment evaluations. These results underscore the effectiveness of integrating explicit cognitive reasoning with reward-based optimization, and highlight it as a promising direction for faithful, preference-aligned multi-role dialogue summarization.

\section{Related Work}
\subsection{Multi-Role Dialogue Summarization}

Multi-role dialogue summarization aims to generate concise and coherent summaries from conversations involving multiple participants. Compared with single-speaker settings, multi-role dialogues pose additional challenges, including intricate turn-taking patterns, rich speaker interactions, and complex discourse structures. Early work in this area primarily relied on extractive strategies, such as graph-based ranking and keyphrase- or template-driven methods, which were widely applied to meeting and multi-party dialogue data~\citep{Murray2005,Riedhammer2008,Tixier2017}. 

With the advent of neural approaches, research has increasingly shifted toward sequence-to-sequence architectures with hierarchical encoders and attention mechanisms to better capture conversational context, discourse structure, and role-specific information~\citep{Li2019,Zhu2020,chen-yang-2020-multi,Feng2021}. More recently, multi-stage and hierarchical frameworks have been proposed to address the challenge of long-range dependencies in extended dialogues and meetings~\citep{Zhang2022}. Nevertheless, faithfully aggregating speaker intent across multi-turn interactions while maintaining factual consistency remains an open challenge, especially in real-world scenarios with diverse and evolving participant roles.

\begin{figure*}[t]
  \centering
  \includegraphics[width=0.9 \textwidth]{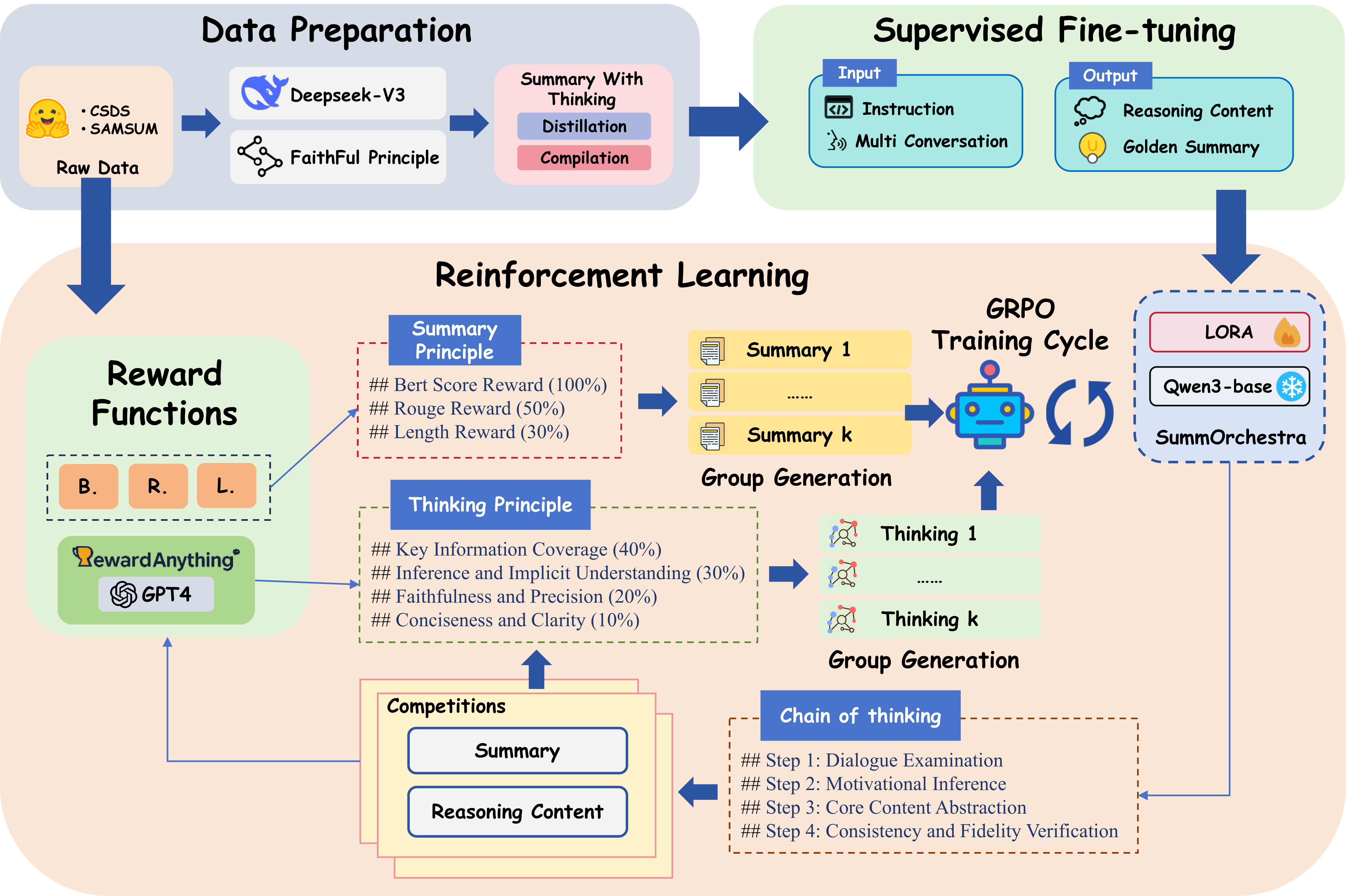}
  \caption{Overview of the proposed framework. The pipeline comprises two main parts: (1) Reasoning distillation from a teacher model to extract thinking traces for data creation; (2) a two-stage training paradigm that first initializes a reasoning-aware summarizer and then internalizes factual consistency and human preferences.}
  \label{fig:method_overview}
\end{figure*}

\subsection{Factual Consistency in Summarization}

In this case, factual consistency has become a central concern in abstractive summarization, where models can generate plausible yet unsupported or contradictory content~\citep{Rennard2023}. To address this, a variety of metrics—such as FactCC~\citep{Kryscinski2019EvaluatingTF}, QAGS~\citep{Feng2024}, SummaC~\citep{Laban2021SummaCRN}, and FIZZ~\citep{Yang2024}—have been introduced to assess the degree to which summaries are grounded in the source. In dialogue summarization, auxiliary language understanding signals have been incorporated to reduce factual errors~\citep{Akani2024}, and discourse-level evaluation highlights the role of document structure, especially in long texts~\citep{Zhong2025}.

Methodologically, prior work includes constrained decoding, post-hoc correction, and fact-checking or entailment modules. For example, graph-based knowledge grounding mitigates unfaithful responses in dialogues~\citep{ji2023rho}. \citet{manakul2023selfcheckgpt} propose SelfCheckGPT to detect hallucinations post-hoc. \citet{yu2024truth} propose Truth-Aware Context Selection (TACS) to filter untruthful context before generation. However, these techniques are mostly designed for single-document or single-speaker settings. Multi-role dialogues pose additional challenges, as facts are fragmented across speakers, expressed informally, and often revised, making robust factual grounding substantially harder.

\subsection{Human Preference Alignment for Text Generation}

Beyond factual correctness, summarization models should align with human preferences for clarity, usefulness, and style. \textit{Reinforcement Learning from Human Feedback (RLHF)} trains models using pairwise human preferences via a learned reward model~\citep{Christiano2017DeepRL}. \textit{Direct Preference Optimization (DPO)} provides a simpler alternative without a separate reward model~\citep{Rafailov2023DirectPO}, with variants such as ODPO~\citep{Amini2024} and length-controlled DPO~\citep{Park2024} addressing preference strength and human bias. 

Recent works further improve factuality and task-specific alignment. Aggregating multiple imperfect metrics enhances factual consistency~\citep{Ye2025b}, while DPO has been applied to conversational recommendation~\citep{Tajiri2025}, multi-dimensional feedback~\citep{Song2025}, and joint instruction-response preference learning (JPO)~\citep{Bansal2025}. Despite these advances, incorporating role-sensitive salience, factuality, and narrative structure into a unified preference signal for multi-role dialogue summarization remains underexplored.

\subsection{Discussion on Limitations of Existing Methods}

Despite notable progress, existing approaches to dialogue summarization exhibit several limitations in multi-role settings. Extractive models struggle to capture implicit relations, discourse-level intent, and cross-speaker dependencies. Abstractive models, while more expressive, are prone to hallucinations, misattribution of roles or responsibilities, and biased emphasis on specific speakers. Moreover, most prior work treats factual consistency and human preference alignment as separate or secondary objectives, and few methods explicitly integrate both in the context of complex, multi-speaker conversations. These gaps motivate the development of frameworks that can simultaneously model role-specific dynamics, enforce factual faithfulness, and optimize for human-aligned quality—precisely the goal of the proposed framework in this work.

\section{Methodology}

\begin{table*}[ht]
  \centering
  \footnotesize
  \resizebox{\linewidth}{!}{
      \begin{tabular}{p{3.5cm} p{9cm} l}
        \toprule
        \textbf{Principle} & \textbf{Description} & \textbf{Weight} \\
        \midrule
        Key Information Coverage &
        Does the summary capture the core facts of the dialogue? Must include: the request/proposal, the refusal, the insistence, and any implied motivation if present. Missing major elements is a critical error. &
        40\% \\
        Inference and Implicit Understanding &
        Does the summary correctly reflect implied attitudes, motives, or emotional tone (e.g., sarcasm, concern, frustration)? Reasonable inference is rewarded. Fabrication is penalized. &
        30\% \\
        Faithfulness and Precision &
        No hallucinations or incorrect claims beyond what can be safely inferred. The summary must not change the meaning of the original dialogue. &
        20\% \\
        Conciseness and Clarity &
        The summary should be brief, readable, and well-structured. Overly verbose summaries lose points even if factually correct. &
        10\% \\
        \bottomrule
      \end{tabular}
  }
  \caption{\label{summary-evaluation-criteria} Evaluation criteria and weighting for dialogue summarization.}
  \label{tab:rewardweighting}
\end{table*}

Our framework for multi-role dialogue summarization is explicitly designed to improve both factual consistency and alignment with human preferences. As illustrated in Figure~\ref{fig:method_overview}, our approach follows a structured pipeline comprising tailored data construction and a two-stage training process, which together equip the model with reasoning-aware and preference-aligned summarization capabilities.

\subsection{Data Construction and Reasoning Distillation}

As shown in the first module of Figure~\ref{fig:method_overview}, we construct high-quality training data for multi-role dialogue summarization by distilling explicit reasoning signals from raw dialogues. Given the complexity of reasoning over multiple speakers and interaction patterns, we employ DeepSeek-v3~\citep{Liu2024} as a teacher model to generate structured reasoning traces—such as step-by-step reasoning paths and intermediate reflections—for each dialogue. These distilled traces are incorporated as auxiliary supervision alongside the original summaries, enabling the student model to internalize both factual content and the underlying reasoning logic that supports faithful summarization and subsequent preference-based training. The distillation process follows the prompt template provided in Appendix~\ref{sec:Distillation Prompt for Multi-role Dialogue Summarization}.

\subsection{Two-Stage Training for Internalizing Faithfulness and Human Preference}

To more effectively internalize reasoning logic and produce summaries that are both faithful and aligned with human preferences, we adopt a two-stage training paradigm. In the first stage, we perform supervised fine-tuning (SFT) to initialize the base model as a reasoning-aware summarizer. The model is initially trained on the original dialogue–summary pairs without reasoning annotations, and is then further fine-tuned on the augmented dataset enriched with distilled reasoning traces from the teacher model. This staged SFT procedure encourages the model to learn the desired reasoning format and leverage intermediate reasoning before generating the final summary, providing a stable and structured initialization for subsequent reinforcement learning.

In the second stage, we apply group relative policy optimization (GRPO) to further align the model’s outputs with our semantic quality and factual principles. For each input dialogue, the policy samples a group of \(G\) candidate summaries, and optimization is carried out based on their relative rewards under the dual-principle reward mechanism. This reinforcement learning phase refines the model beyond supervised imitation, driving it toward summaries that are more factually faithful and better reflect human preferences in complex multi-role conversational settings. The group-level reward is defined in Eq.~\ref{eq:groupReward}:
\begin{equation}
\begin{aligned}
R_\text{group} 
&= \frac{1}{G} \sum_{i=1}^{G} R_i^\text{base}, \\[1mm]
R_i^\text{base} 
&= \lambda_b \, R_{\text{bertscore},i} 
  + \lambda_r \, R_{\text{rouge},i} \\
&\quad + \lambda_l \, R_{\text{length},i} 
  + \lambda_p \, R_{\text{principle},i}
\end{aligned},
\label{eq:groupReward}
\end{equation}
where $\lambda_b : \lambda_r : \lambda_l : \lambda_p = 1 : 0.5 : 0.3 : 1$.

We design four complementary reward components to guide GRPO training. $R_{\text{rouge}}$ is defined as the average of ROUGE-1, ROUGE-2, and ROUGE-L, encouraging reasonable lexical overlap with the reference without overfitting to surface forms. $R_{\text{bertscore}}$ measures semantic similarity via contextual embeddings, promoting preservation of sentence-level meaning. $R_{\text{length}}$ rewards summaries whose length is close to that of the reference, discouraging both under- and over-generated outputs. Finally, $R_{\text{principle}}$ leverages the RewardAnything framework~\citep{Yu2025} with GPT-4 as a teacher model to score summaries against four human-aligned criteria—key information coverage, implicit inference, factual faithfulness, and conciseness (Table~\ref{summary-evaluation-criteria}).

We set the reward weights, as shown in Table~\ref{tab:rewardweighting}, to emphasize semantic consistency and factual correctness while preserving concise expression. In particular, the ROUGE-based component is assigned a relatively lower weight to avoid mere imitation of reference phrasing, whereas the principle-based reward plays a dominant role in steering the model toward factually grounded, preference-aligned summaries in multi-role dialogue settings.

\section{Experimental Results}
\subsection{Experimental Setup}

\paragraph{Benchmark Details.} 
We conduct experiments on two widely used dialogue summarization benchmarks: CSDS~\citep{Lin2021} and SAMSum~\citep{Gliwa2019}. CSDS is a large-scale Chinese dialogue summarization dataset constructed from multi-turn, multi-party conversations in diverse, real-world scenarios (e.g., customer service, daily chat, and task-oriented discussions). It provides human-written abstractive summaries for each dialogue, with conversations typically longer and more complex than in standard English benchmarks. This makes CSDS particularly suitable for evaluating models’ ability to handle informal language, topic drift, and information sparsity in Chinese conversational settings.

SAMSum is an English dialogue summarization dataset composed of short, messenger-style conversations that resemble everyday chat on platforms such as WhatsApp or Facebook Messenger. Each dialogue is paired with a concise abstractive summary written by linguists, focusing on capturing key events, decisions, and user intentions. Compared to CSDS, SAMSum dialogues are generally shorter and more informal, with a strong emphasis on ellipsis, pragmatics, and conversational implicature, thus providing a complementary testbed for evaluating cross-lingual robustness and performance on casual English conversations.

\paragraph{Evaluation Metrics.}
We assess models with complementary metrics covering lexical overlap, semantic similarity, factual consistency, and human preference, using language-aware configurations for multilingual data. Together, these metrics provide a compact yet comprehensive evaluation of multi-role dialogue summaries, balancing surface overlap, meaning preservation, factual grounding, and alignment with human preferences.

We report ROUGE-1, ROUGE-2, and ROUGE-L with language-specific toolchains \cite{lin-2004-rouge}. Language is detected via a reference-based classifier (langid \footnote{\url{https://github.com/saffsd/langid.py}}). For Chinese, we use rouge\_chinese \footnote{\url{https://github.com/Isaac-JL-Chen/rouge_chinese}} with THULAC \cite{li2009punctuation} word-level tokenization; for English, we use py-rouge \footnote{\url{https://github.com/diegoantognini/py-rouge}}. This avoids artifacts from applying a single, non-adapted ROUGE variant across languages. Details are present in Table~\ref{tab:rouge_protocol}.

\begin{table}[ht]
  \centering
  \small
  \resizebox{\linewidth}{!}{
      \begin{tabular}{ll}
        \toprule
        \textbf{ROUGE Setting} & \textbf{Description} \\
        \midrule
        Language detection & langid (reference-based) \\
        Chinese ROUGE & \texttt{rouge\_chinese} \\
        Chinese tokenization & THULAC (word-level) \\
        Token granularity & Word-level (not character-level) \\
        Character substitution & Not used \\
        English ROUGE & \texttt{py-rouge} \\
        Metrics & ROUGE-1 / ROUGE-2 / ROUGE-L \\
        \bottomrule
      \end{tabular}
  }
  \caption{Evaluation protocol for ROUGE.}
  \label{tab:rouge_protocol}
\end{table}

Semantic similarity is measured with BERTScore \cite{zhang2019bertscore} using language-specific backbones: BERT-base Chinese \footnote{\url{https://huggingface.co/google-bert/bert-base-chinese}} for Chinese texts and BERT-base uncased \footnote{\url{https://huggingface.co/google-bert/bert-base-uncased}} for English. Scores are computed with cosine similarity over the top 12 layers, and we report mean precision, recall, and F1, without IDF weighting. Details are present in Table~\ref{tab:bertscore_protocol}.

\begin{table}[ht]
  \centering
  \small
  \resizebox{\linewidth}{!}{
      \begin{tabular}{ll}
        \toprule
        \textbf{BERTScore Setting} & \textbf{Description} \\
        \midrule
        Language detection & langid (reference-based) \\
        Chinese backbone & BERT-base Chinese \\
        English backbone & BERT-base uncased \\
        Layer selection & Top 12 transformer layers \\
        Score aggregation & Mean Precision / Recall / F1 \\
        IDF weighting & Not used \\
        Token alignment & Cosine similarity \\
        \bottomrule
      \end{tabular}
  }
  \caption{Evaluation protocol for BERTScore.}
  \label{tab:bertscore_protocol}
\end{table}

Factual consistency is quantified as the proportion of a summary’s factual claims that are supported by the source dialogue. We first decompose each summary into atomic statements, then verify each against the dialogue using HHEM-2.1-Open~\citep{hhem-2.1-open}, a T5-based hallucination detector. The faithfulness score is: $\text{Faithfulness} = \frac{\# supported\ claims}{\# total\ claims}$, yielding a value in $[0,1]$.

We adopt WorldPM-72B-RLHFLOW~\citep{wang2025worldpm} as an automated evaluator to approximate human preference judgments over model outputs. WorldPM has been extensively validated and shown to match or surpass the reliability of human annotators across a wide range of evaluation tasks~\citep{wang2025worldpm}, providing a strong empirical basis for its use as a proxy for human evaluation. Leveraging WorldPM enables us to systematically and scalably measure preference signals without relying exclusively on expensive, time-consuming human annotation. This design choice is consistent with emerging best practices in alignment research, where high-capacity reward models and LLM-as-judge frameworks are increasingly used to approximate human feedback in both RLHF pipelines and automatic preference evaluation~\citep{li-etal-2025-generation,Wu2025RewardDanceRS}.

\begin{figure*}[htbp]
    \centering
    \begin{subfigure}{0.48\textwidth}
        \centering
        \includegraphics[width=\linewidth]{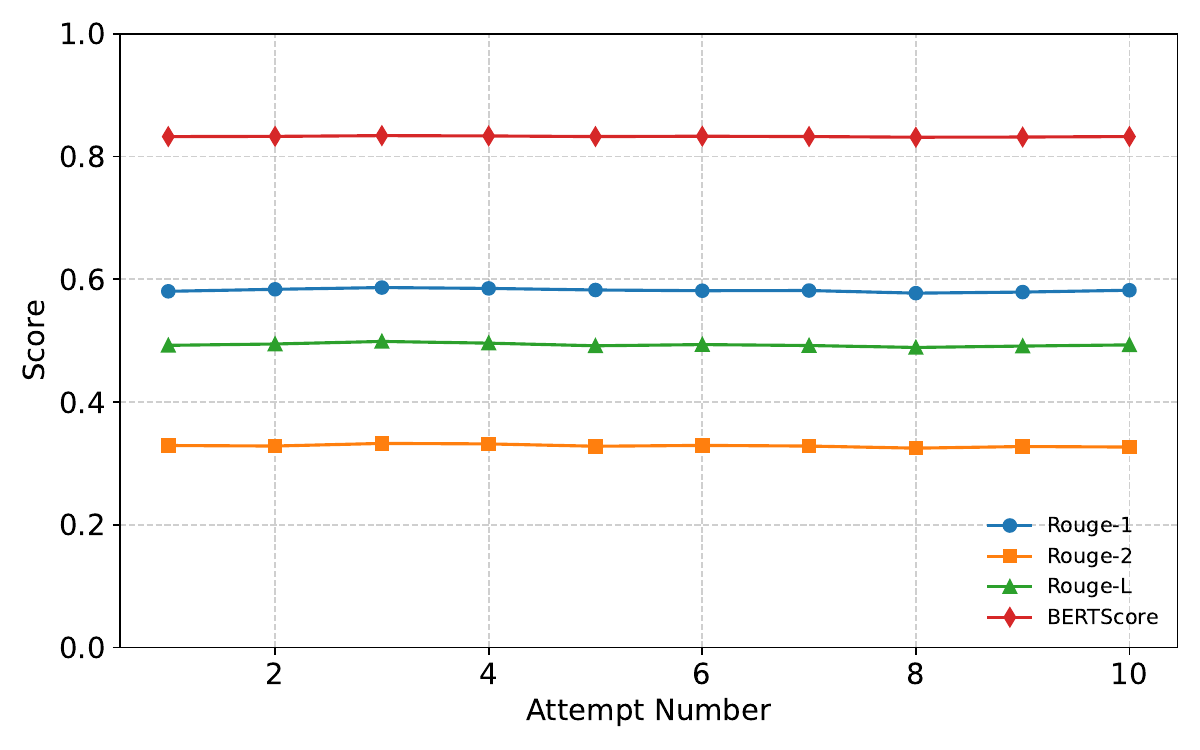}
        \caption{CSDS dataset}
        \label{fig:csds_sampling}
    \end{subfigure}
    \hfill
    \begin{subfigure}{0.48\textwidth}
        \centering
        \includegraphics[width=\linewidth]{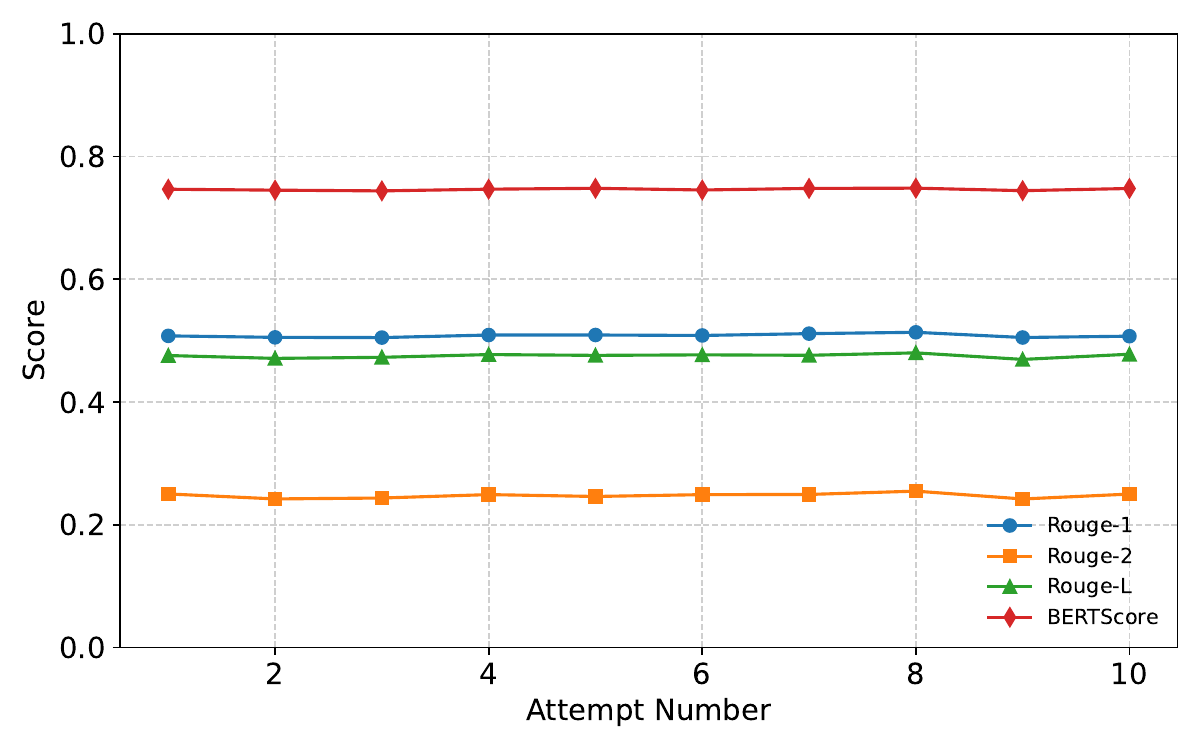}
        \caption{SAMSum dataset}
        \label{fig:samsum_sampling}
    \end{subfigure}
    \captionsetup{skip=2pt}
    \caption{Performance of our model over ten resampling trials on the test datasets of CSDS and SAMSum}
    \label{fig:csds_samsum_sampling}
\end{figure*}

\paragraph{Training Detail.} 
The SFT stage employs a LoRA-based parameter-efficient training strategy. Models are trained for 8 epochs with a per-device batch size of 16, learning rate $1 \times 10^{-5}$, and gradient accumulation of 1 step. Adam optimizer is applied to all linear modules. Maximum sequence length is 2048, with mixed precision and gradient checkpointing enabled. Validation and checkpointing occur every 50 steps, with early stopping after 3 intervals without improvement. LoRA parameters are set to rank 16 and alpha 32, with 8 dataloader workers and a warmup ratio of 0.05.  

Following SFT, the GRPO stage further refines the model using a training reward composed of four components: ROUGE, BERTScore, length, and a \textbf{principle reward} implemented via RewardAnything. GRPO training is conducted for 100 epochs with a per-device batch size of 12, learning rate $5 \times 10^{-6}$, maximum completion length of 2048, gradient clipping of 0.5, and mixed precision (bfloat16). Dataset processing uses 8 processes with 4 generations per example. Reward ablation experiments are conducted on subsets of the four reward components: B (BERTScore), R (ROUGE), L (length), and P (principle).

\paragraph{Evaluation Setting.} 
For evaluation, we measure \textbf{faithfulness}, ensuring factual consistency with source dialogues and distilled reasoning, and \textbf{human preference}, assessing alignment with human judgments of clarity, informativeness, and relevance. This distinction clarifies that the training rewards and evaluation metrics are not identical.

All experiments are conducted on 8 Ascend NPU 910B cards, each with 64GB of memory. We evaluate Qwen2.5 on CSDS and Qwen3 on SAMSum, selecting the base models that best align with each dataset's language characteristics and reasoning requirements (see Appendix~\ref{app:trainingDetails}). Training follows a two-stage paradigm: supervised fine-tuning (SFT) followed by guided reinforcement learning with preferences (GRPO). More training details can be found in Appendix~\ref{app:trainingDetails}. For reproducibility, all model and datasets are publicly available.\footnote{\url{https://huggingface.co/collections/NebulaPixel/summorchestra-multirole-summary}}

\subsection{Experimental Results Analysis}

\begin{table*}[!h]
  \centering
  \resizebox{\linewidth}{!}{
      \begin{tabular}{llcccccccc}
        \hline
        \textbf{Model} & \textbf{Training Method} & \textbf{B} & \textbf{R} & \textbf{L} & \textbf{P} & \textbf{ROUGE-1} & \textbf{ROUGE-2} & \textbf{ROUGE-L} & \textbf{BERTScore} \\
        \hline
        gpt-4.1      & -               & - & - & - & - & 42.91 & 14.08 & 32.04 & 77.37 \\
        gpt-4.1-mini & -               & - & - & - & - & 42.78 & 14.11 & 32.31 & 77.40 \\
        gpt-4o       & -               & - & - & - & - & 48.52 & 19.59 & 36.55 & 79.51 \\
        gpt-4o-mini  & -               & - & - & - & - & 45.87 & 16.93 & 33.82 & 78.28 \\
        gpt-5        & -               & - & - & - & - & 39.72 & 11.47 & 28.98 & 75.63 \\
        gpt-5-mini   & -               & - & - & - & - & 41.54 & 12.95 & 30.15 & 76.68 \\
        \hline
        qwen2.5-3B   & SFT             & - & - & - & - & 50.87 & 27.35 & 43.40 & 76.58 \\
        qwen2.5-3B   & GRPO(B)         & \checkmark & - & - & - & 47.21 & 25.78 & 41.85 & 83.10 \\
        qwen2.5-3B   & GRPO(B+R)       & \checkmark & \checkmark & - & - & 58.00 & 32.95 & 48.90 & 83.22 \\
        qwen2.5-3B   & GRPO(B+R+L)     & \checkmark & \checkmark & \checkmark & - & 58.50 & 33.25 & 49.20 & 83.28 \\
        \hline
        qwen2.5-7B   & SFT             & - & - & - & - & 51.11 & 28.19 & 44.05 & 77.29 \\
        qwen2.5-7B   & GRPO(B)         & \checkmark & - & - & - & 48.40 & 23.45 & 42.35 & 83.42 \\
        qwen2.5-7B   & GRPO(B+R)       & \checkmark & \checkmark & - & - & 58.80 & 34.00 & 49.75 & 83.50 \\
        qwen2.5-7B   & GRPO(B+R+L)     & \checkmark & \checkmark & \checkmark & - & 59.05 & 34.25 & 50.05 & 83.58 \\
        \hline
      \end{tabular}
  }
  \caption{\label{tab:csds_results}
    Results on the CSDS dataset. B, R, L represent the rewards enabled in GRPO training: 
    B for BERTScore, R for ROUGE, L for length reward.
    Checkmarks indicate which rewards were used in each training method.
  }
\end{table*}

\begin{table*}[!h]
  \centering
  \begin{threeparttable}
  \resizebox{\linewidth}{!}{
      \begin{tabular}{lcccccc}
        \hline
        \textbf{Model} &
        \textbf{ROUGE-1} & \textbf{ROUGE-2} & \textbf{ROUGE-L} &
        \textbf{BERTScore} & \textbf{Aligned Pref.} & \textbf{Faithfulness} \\
        \hline
        Original Dataset & -- & -- & -- & -- & 0.4973 & 0.7548 \\
        \hline
        gpt-4.1        & 42.91 & 14.08 & 32.04 & 77.37 & 0.5012 & 0.6124 \\
        gpt-4.1-mini   & 42.78 & 14.11 & 32.31 & 77.40 & 0.4983 & 0.6021 \\
        gpt-4o         & 48.52 & 19.59 & 36.55 & 79.51 & 0.5223 & 0.6354 \\
        gpt-4o-mini    & 45.87 & 16.93 & 33.82 & 78.28 & 0.5434 & 0.6132 \\
        gpt-5          & 39.72 & 11.47 & 28.98 & 75.63 & 0.5635 & 0.6144 \\
        gpt-5-mini     & 41.54 & 12.95 & 30.15 & 76.68 & 0.4832 & 0.5344 \\
        \hline
        qwen3-fast\tnote{a} (BRL)     & 48.85 & 22.89 & 45.86 & 72.64 & 0.4435 & 0.6659 \\
        qwen3-fast\tnote{a} (BRLP)    & 47.40 & 21.79 & 44.73 & 70.34 & \textbf{0.5563} & \textbf{0.7959} \\
        qwen3-slow\tnote{b} (BRL) 
                       & 50.78 & 25.03 & 47.58
                       & 74.67 & 0.4883 & 0.6852 \\
        qwen3-slow\tnote{b} (BRLP) 
                       & 49.78 & 24.12 & 46.43
                       & 73.32 & \textbf{0.6683} & \textbf{0.8352} \\
        \hline
      \end{tabular}}
      \begin{tablenotes}
        \footnotesize
        \item[a] ``fast'' denotes the no-thinking (fast inference) mode.
        \item[b] ``slow'' denotes the thinking (slow inference) mode.
      \end{tablenotes}
      \caption{\label{tab:samsum_results}
        Results on the SAMSum dataset.
        For qwen3, both modes are evaluated after supervised fine-tuning and GRPO training.
        \textbf{The principle reward (P) enforces factual consistency and structural validity, and does not directly optimize ROUGE or BERTScore.}
    }
\end{threeparttable}
\end{table*}

To ensure the reliability of our evaluation, we first assessed the stability of the trained models on the CSDS and SAMSum test sets using ten resampling trials. As shown in Figure~\ref{fig:csds_samsum_sampling}, the results remain consistently stable across these trials, indicating that the final models generalize well and do not exhibit signs of overfitting. 

Building upon this, Table~\ref{tab:csds_results} summarizes the performance of various models on the CSDS dataset. It can be seen that supervised fine-tuning (SFT) with Qwen2.5 models leads to substantial improvements over baseline LLMs. Furthermore, the GRPO stage with semantic and lexical rewards (B+R+L) consistently improves over SFT, demonstrating the framework's effectiveness in optimizing standard metrics on large-scale data.

During the GRPO stage, the choice of reward signals significantly influences model behavior. Optimizing solely for semantic similarity tends to reduce surface-level overlap, while incorporating additional rewards, such as ROUGE and length, progressively enhances both semantic and structural quality. These observations suggest that different reward components contribute complementary benefits, and a careful combination is crucial for balanced performance.

\subsection{Ablation Study}

To further dissect the effects of reward design, we analyze results on the SAMSum dataset, comparing fast and slow inference regimes for Qwen3 models. Incorporating the principle reward leads to modest reductions in conventional metrics but substantially improves factual faithfulness and preference alignment. As illustrated in Figure~\ref{fig:samsum_human_preference_faithfulness}, the principle reward markedly enhances the preference score and factual reliability, demonstrating that structured rewards can guide models toward outputs that better align with human preference criteria. Notably, our model (0.6683) surpasses the original human references (0.4973) in preference scores, suggesting it generates more coherent and useful summaries than the noisy ground truth.

\subsection{Summary of Findings}

\begin{figure}[ht]
    \centering
    \includegraphics[width=\linewidth]{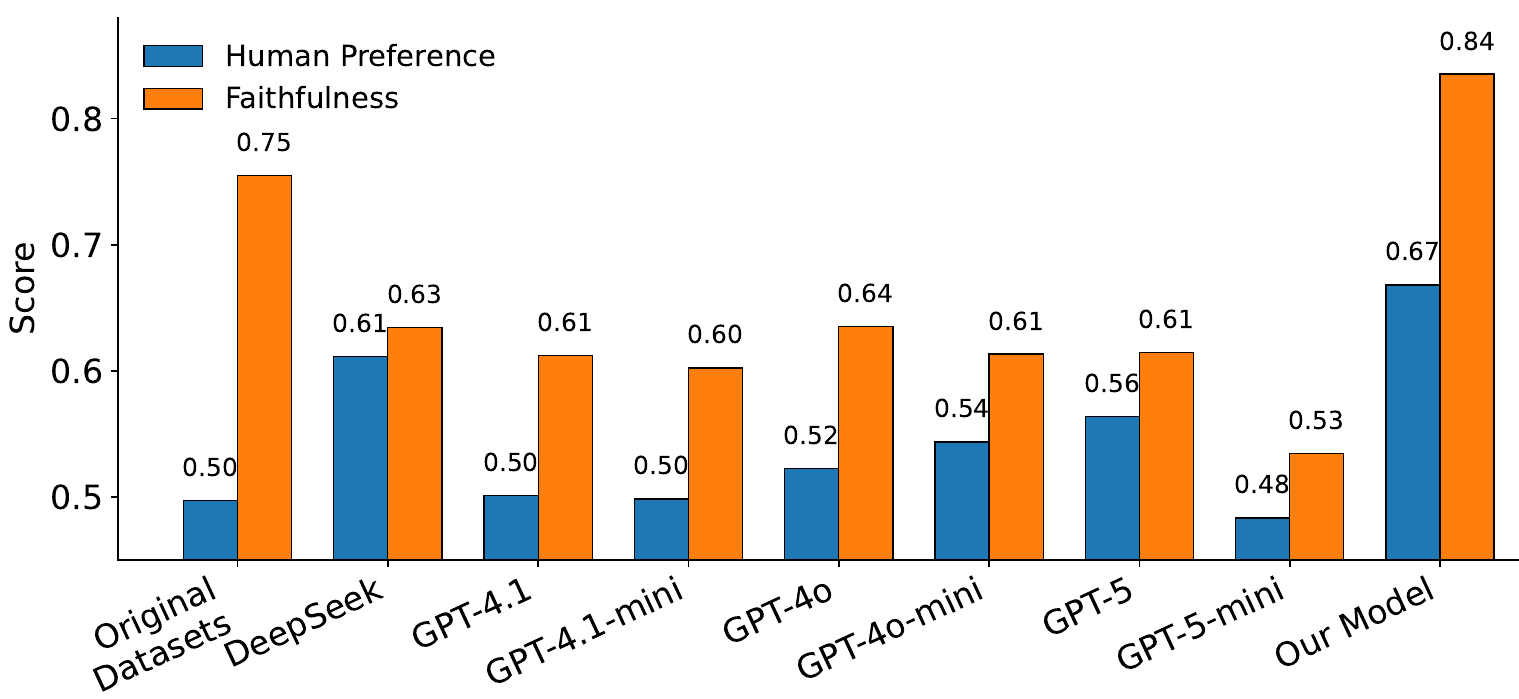}
    \caption{Human Preference and Faithfulness on the SAMSUM Dataset}
    \label{fig:samsum_human_preference_faithfulness}
\end{figure}

The above experiments reveal a consistent pattern: traditional automatic metrics alone do not fully capture summary quality, particularly regarding factual consistency and user-perceived value. Multi-reward GRPO training improves both semantic and structural aspects of summaries, while the addition of the principle reward shifts optimization toward structural validity and factual reliability. The improvements in human preference and faithfulness, corroborated by Figure~\ref{fig:samsum_human_preference_faithfulness}, underscore the importance of integrating both surface-level and principle-oriented objectives when optimizing models for multi-role dialogue summarization, providing a holistic measure of model performance.

\section{Conclusion and Future Work}

We presented a two-stage framework for multi-role dialogue summarization that explicitly targets factual consistency and alignment with human preferences rather than merely optimizing surface-level metrics. First, a teacher–student SFT stage distills structured cognitive-style reasoning into a reasoning-aware summarizer. Then, a GRPO-based reinforcement learning stage applies a dual-principle reward that jointly supervises intermediate reasoning traces and final summaries along dimensions of key information coverage, implicit inference, factual faithfulness, and conciseness.

Experiments on CSDS and SAMSum show that our approach matches strong baselines on ROUGE and BERTScore while achieving clear gains in factual faithfulness and model-based preference alignment, supported by qualitative evidence of more accurate and coherent summaries. These results underscore the limitations of traditional metrics for multi-role dialogue summarization and highlight the value of explicitly integrating reasoning and preference-aware objectives into training.

Future work includes constructing datasets that more directly capture factual consistency and human preferences, scaling our framework to larger and more diverse multilingual dialogue corpora, and exploring broader applications of reasoning-aware, preference-aligned optimization beyond summarization.

\section*{Limitations}

While our two-stage framework demonstrates strong performance in multi-role dialogue summarization, several limitations remain.

First, evaluation on the Chinese CSDS dataset is constrained by the lack of suitable factual-consistency and preference evaluators. Because HHEM-2.1-Open cannot be applied to Chinese dialogues, the GRPO stage for CSDS omits the principle reward, limiting our ability to directly optimize and assess faithfulness on this dataset. In addition, SAMSum is relatively small, predominantly English, and exhibits limited topical diversity, which may restrict the generalization of our approach to more complex, multilingual, or domain-specific multi-role dialogues.

Second, our GRPO reward function combines automatic metrics (ROUGE, BERTScore) with model-based proxies for factual consistency and human preference. Although this design captures several aspects of summary quality, the available human preference annotations are limited in scale and potentially subjective, and existing factual-consistency metrics may miss subtle hallucinations or nuanced errors.

While intermediate reflections improve summarization performance on the evaluated benchmarks, their effectiveness across other domains, languages, and more complex dialogue settings remains to be validated. Additionally, reflection introduces extra training and inference overhead, posing challenges for large-scale deployment.

Future work will focus on expanding and diversifying datasets, developing more reliable evaluators (especially for non-English settings), and improving training and inference efficiency to enhance robustness, generalization, and real-world applicability.
\newpage

\section*{Acknowledgments}
This work was supported in part by the Research Project on Ideological and Political Work under the Philosophy and Social Science Planning Program of Zhejiang Province (No.~25GXSZ009YB), and in part by the National Key Research and Development Program of China (No.~2022YFC3303600).

\bibliography{AAA-reference}

\newpage
\appendix

\section{Training Details}
\label{app:trainingDetails}

This appendix provides detailed descriptions of the dataset-specific training strategies
and the two-stage optimization procedure adopted in our framework.

\subsection{Dataset-Specific Training Strategies}

We adopt tailored training strategies for different datasets to accommodate their
language characteristics and annotation availability. The dataset statistics and
splitting strategies are also summarized for completeness.

\begin{itemize}
    \item \textbf{CSDS (Chinese)}:
    The CSDS dataset contains 9,101 training samples, 800 validation samples, and
    800 test samples. For the Chinese multi-role dialogue dataset CSDS, we employ
    the fast-thinking model Qwen2.5. Since CSDS does not provide reliable intermediate
    reasoning annotations and automatic faithfulness evaluators are not readily
    available for Chinese, we directly train the model using supervised fine-tuning
    (SFT) on dialogue--summary pairs.

    Specifically, SFT is performed on the union of the training and validation sets,
    with the validation split also used for model selection. For the reinforcement
    learning stage, we further utilize the full dataset, and reserve the first 200
    samples from the validation set as the evaluation set during GRPO training.
    Final performance is reported on the held-out test set.

    \item \textbf{SAMSum (English)}:
    The SAMSum dataset consists of 14,700 training samples, 818 validation samples,
    and 819 test samples. For the English SAMSum dataset, we use the slow-thinking
    model Qwen3, which supports explicit reasoning representations.

    We first perform knowledge distillation to augment the original dataset with
    distilled intermediate reasoning signals. The augmented data are then used for
    supervised fine-tuning, where both the training and validation sets are jointly
    used for training and model selection. Subsequently, guided reinforcement learning
    with preferences (GRPO) is applied on the full dataset, with the first 200 samples
    from the validation set reserved as the evaluation set during training.

    Final evaluation is conducted on the test set to ensure fair comparison.
\end{itemize}

\subsection{Two-Stage Training for Internalizing Faithfulness and Human Preference}

We employ a two-stage optimization pipeline consisting of supervised fine-tuning
and preference-guided reinforcement learning.

\paragraph{Supervised Fine-Tuning.}
We first initialize the base models via supervised fine-tuning.
For Qwen2.5, we directly fine-tune the model on the CSDS dataset using standard
dialogue--summary supervision.
For Qwen3, we adopt a two-stage SFT strategy on the SAMSum dataset:
the model is first fine-tuned on the original data without reasoning annotations,
and subsequently fine-tuned on an augmented version containing distilled reasoning
signals. This procedure provides a stable initialization for reinforcement learning.

\paragraph{Guided Reinforcement Learning with Preferences.}
After supervised fine-tuning, we further optimize the model using
Guided Reinforcement Learning with Preferences (GRPO).
For each input dialogue, the model samples a group of $G$ candidate summaries,
and optimization is performed based on their relative rewards.
This stage encourages the model to generate summaries that are not only
semantically informative but also factually consistent and aligned with
human judgment.

\section{Evaluation Metrics Details and Modification}
\label{app:evalMetric}
We evaluate model performance using automatic metrics with language-aware configurations to ensure fair and accurate assessment across multilingual dialogue datasets.
Unlike prior work that applies a unified evaluation toolkit regardless of language, we explicitly distinguish between Chinese and English inputs and adopt language-specific implementations for both ROUGE and BERTScore.

\paragraph{ROUGE.}

As summarized in Table~\ref{tab:rouge_protocol}, we first identify the dialogue language using reference-based language detection.
For Chinese samples, we apply the \texttt{rouge\_chinese} toolkit with THULAC-based word-level tokenization, rather than character-level segmentation.
For English samples, we use the standard \texttt{py-rouge} implementation.
In both cases, we report ROUGE-1, ROUGE-2, and ROUGE-L scores.
This design avoids inaccuracies caused by applying non-adapted ROUGE variants to languages with different tokenization characteristics.

\paragraph{BERTScore.}

BERTScore is computed following the protocol in Table~\ref{tab:bertscore_protocol}, with language-specific backbone models.
We use BERT-base Chinese for Chinese texts and BERT-base uncased for English texts, and aggregate similarity scores across the top 12 transformer layers.
Scores are computed using cosine similarity without IDF weighting, and we report the mean precision, recall, and F1 scores.
Language detection is performed consistently with the ROUGE evaluation to ensure aligned metric computation.

\paragraph{Faithfulness.}

In addition to lexical and semantic similarity metrics, we evaluate factual faithfulness to assess whether the generated summaries are consistent with the source dialogue content.
Faithfulness measures the degree to which the factual statements in a model-generated summary are supported by the corresponding dialogue context, with scores ranging from 0 to 1, where higher values indicate better factual consistency.

Formally, a summary is considered faithful if all of its factual claims can be inferred from the source dialogue.
The faithfulness score is computed as the proportion of supported claims among all identified claims in the summary:
\begin{equation}
\text{Faithfulness} = 
\frac{\text{\# supported claims}}{\text{\# total claims in the summary}}.
\end{equation}

To operationalize this metric in the multi-role, multi-turn dialogue summarization setting, we first decompose each generated summary into a set of atomic factual statements.
Each statement is then individually verified against the original dialogue context.
For this verification step, we employ \textbf{HHEM-2.1-Open}~\citep{hhem-2.1-open}, a T5-based classifier model released by Vectara, which is specifically trained to detect hallucinations in LLM-generated text.
Given a candidate statement and the dialogue context, the model predicts whether the statement can be inferred from the context.
The final faithfulness score is obtained by averaging the binary verification results over all statements in the summary.

\paragraph{Human Preference.}

To further assess summary quality beyond automatic similarity and factual consistency metrics, we incorporate a human preference evaluation.
This metric aims to capture holistic quality aspects that are difficult to quantify automatically, including informativeness, coherence, readability, and overall usefulness of the summary.

Human preference scores are computed using the \textbf{WorldPM-72B-RLHFLOW} \cite{wang2025worldpm} model, a large-scale preference model trained on extensive human feedback data.
The training corpus of WorldPM-72B-RLHFLOW includes a wide range of instruction-following and summarization tasks, enabling it to reliably approximate human judgments in dialogue summarization scenarios.
Given a pair consisting of the source dialogue and a candidate summary, the model produces a scalar preference score reflecting the likelihood that a human annotator would favor the summary.

We report the average human preference score across all evaluation samples.
By combining human preference evaluation with ROUGE, BERTScore, and faithfulness metrics, we provide a more comprehensive assessment of summary quality, especially in multi-role dialogue settings where factual accuracy and overall coherence are both critical.

Overall, this language-aware evaluation protocol provides a more reliable comparison across multilingual dialogue summaries and reduces potential bias introduced by mismatched evaluation tools.

\section{Distillation Prompt for Multi-role Dialogue Summarization}
\label{sec:Distillation Prompt for Multi-role Dialogue Summarization}

\begin{tcolorbox}[
  colback=blue!5!white,
  colframe=blue!70!black,
  title=Multi-role Dialogue Summarization Distillation Prompt,
  breakable
]
You are a large language model participating in a \emph{knowledge distillation} process
for multi-role, multi-turn dialogue summarization.

\medskip
\textbf{Task Definition.}  
Given a dialogue and its corresponding \emph{reference summary} (generated by a stronger teacher model or human annotators),
your goal is to reproduce a high-quality summary by explicitly reasoning over the dialogue content.
The reference summary is provided \emph{only as supervision} and should not be copied verbatim.

\medskip
\textbf{Reasoning Requirement.}  
Before producing the final summary, you must explicitly perform structured reasoning
by outputting a \texttt{<think></think>} block that follows the four steps below:

\begin{enumerate}
    \item \textbf{Dialogue Examination}:  \\
    Identify speakers, dialogue turns, and key utterances across the multi-role interaction.
    \item \textbf{Motivational Inference}:  \\
    Infer speakers' intentions, preferences, or implicit goals behind their statements.
    \item \textbf{Core Content Abstraction}:  \\
    Extract and abstract the essential information, including requests, responses, agreements, or conflicts.
    \item \textbf{Consistency and Fidelity Verification}:  \\
    Verify that the abstracted content is faithful to the original dialogue and aligned with the reference summary,
    avoiding omissions, distortions, or hallucinated information.
\end{enumerate}

\medskip
\textbf{Example Input:}

\begin{lstlisting}[
  basicstyle=\ttfamily\small,
  breaklines=true,
  breakatwhitespace=true
]

Conversation:
Amanda: I baked cookies. Do you want some? 
Jerry: Sure! Amanda: I'll bring you tomorrow :-)

Summary:
Amanda baked cookies and will bring Jerry some tomorrow.
\end{lstlisting}

\textbf{Example Output:}

\begin{lstlisting}[
  basicstyle=\ttfamily\small,
  breaklines=true,
  breakatwhitespace=true
]
<think>I observed Amanda stating she baked cookies and offering some to Jerry, who accepted, followed by Amanda confirming she will bring them tomorrow; this sequence shows a clear offer, acceptance, and planned delivery, leading me to conclude Amanda baked cookies and will share them with Jerry the next day.</think>
Amanda baked cookies and will bring Jerry some tomorrow.
\end{lstlisting}

\medskip
\textbf{Note.}  
This prompt is used exclusively during the \emph{distillation stage} to enable the student model
to learn structured reasoning behaviors when trained on the SAMSum dataset,
whose original annotations contain summaries without explicit thinking or reasoning traces.
\end{tcolorbox}

\section{Summary Reward Principle}
\label{sec:Summary Reward Principle}

\begin{tcolorbox}[
  colback=gray!5,
  colframe=gray!40,
  title=Summary Reward Principle,
  boxrule=0.5pt,
  left=6pt,
  right=6pt,
  top=6pt,
  bottom=6pt,
]
\setlength{\parskip}{0.6em}
\setlength{\parindent}{0pt}

Evaluate summaries using these criteria:

1. \textbf{Key Information Coverage (40\%)}  
- Does the summary capture the core facts of the dialogue?  
- Must include: the request/proposal, the refusal, the insistence,
  and any implied motivation if present.  
- Missing major elements is a critical error.

2. \textbf{Inference and Implicit Understanding (30\%)}  
- Does the summary correctly reflect implied attitudes, motives,
  or emotional tone (e.g., sarcasm, concern, frustration)?  
- Reasonable inference is rewarded. Fabrication is penalized.

3. \textbf{Faithfulness and Precision (20\%)}  
- No hallucinations or incorrect claims beyond what can be safely inferred.  
- The summary must not change the meaning of the original dialogue.

4. \textbf{Conciseness and Clarity (10\%)}  
- The summary should be brief, readable, and well-structured.  
- Overly verbose summaries lose points even if factually correct.

\medskip
\textbf{Priority:}  
Key information coverage $>$ faithfulness $>$ inference quality $>$ clarity.
\end{tcolorbox}

\vfill\eject
\section{Principle-based Scorer Prompt Template}
\label{appendix:principle_eval_prompt}

This appendix provides the prompt template used for principle-based evaluation. The evaluator is instructed to assess model outputs strictly according to a predefined evaluation principle.

\begin{tcolorbox}[
  colback=gray!5,
  colframe=gray!60,
  title=\textbf{System Prompt for Principle-based Scorer},
  boxrule=0.5pt,
  arc=2pt,
  left=6pt,
  right=6pt,
  top=6pt,
  bottom=6pt
]
\small

\textbf{SYSTEM PROMPT}

You are an evaluator judging model responses based on a given \textbf{evaluation principle}. Your primary goal is to assess how well each response adheres to the principle, prioritizing this over general preferences, while avoiding endorsement of harmful content.

\medskip
\textbf{Evaluation Procedure:}
\begin{enumerate}
  \item Carefully read the principle, the input prompt, and all candidate responses. Briefly consider how each response aligns with the principle in a concise reasoning process.
  
  \item Assign a score to each response on a 1--5 scale:
  \begin{itemize}
    \item 5: Perfect adherence with excellent overall quality
    \item 4: Strong adherence with minor limitations
    \item 3: Basic adherence
    \item 2: Partial adherence with important omissions
    \item 1: Poor adherence or contradiction to the principle
  \end{itemize}
  
  \item Rank all responses from best to worst. Responses with identical scores should still be ordered based on relative quality.
\end{enumerate}

\medskip
Use the full scoring range when necessary. Avoid score compression if there are clear quality differences among responses.

\medskip
\textbf{Output Format (JSON only):}
\begin{verbatim}
{
  "scores": {"model-1": 2, "model-2": 4, ...},
  "best-to-worst": ["model-2", "model-1", ...]
}
\end{verbatim}

\end{tcolorbox}

\end{sloppypar}
\end{document}